\documentclass[11pt,a4paper]{article}
\usepackage[hyperref]{eacl2021}
\usepackage{times}
\usepackage{latexsym}

\usepackage{comment}
\usepackage{hyperref}
\usepackage{dingbat}
\usepackage{multirow}
\usepackage{enumitem}
\usepackage{caption}
\usepackage{subcaption}
\usepackage{adjustbox,booktabs}
\usepackage{tikz}
\def\checkmark{\tikz\fill[scale=0.4](0,.35) -- (.25,0) -- (1,.7) -- (.25,.15) -- cycle;}

\usepackage{microtype}

\aclfinalcopy 
\def\aclpaperid{40}

\title{\ds: Arabic COVID-19 Twitter Dataset for Misinformation Detection
}

\author{Fatima Haouari, Maram Hasanain, Reem Suwaileh, Tamer Elsayed \\
  Computer Science and Engineering Department, Qatar University \\
  \texttt{\{200159617, maram.hasanain, rs081123, telsayed\}@qu.edu.qa}}

\date{}
\usepackage{soul} 
\definecolor{lightblue}{rgb}{.50,.90,0.51}
\definecolor{tri}{rgb}{.25,.88,.82}
\definecolor{lilac}{rgb}{0.85,0.64,0.85}
\definecolor{atomictangerine}{rgb}{1.0, 0.6, 0.4}
\newcommand{\hlc}[2][yellow]{ {\sethlcolor{#1} \hl{#2}} }
\newcommand\fh[1]{\hlc[lightblue]{{\bf FH}: #1}}
\newcommand\rs[1]{\hlc[lilac]{{\bf RS}: #1}}

\newcommand{\dscore}{ArCOV-19}
\newcommand{\ds}{ArCOV19-\emph{Rumors}}
\usepackage{graphicx}
\usepackage{float}

\begin{document}

\maketitle
\begin{abstract}
In this paper we introduce \ds, an Arabic COVID-19 Twitter dataset for misinformation detection composed of tweets containing claims from 27$^{th}$ January till the end of April 2020. We collected 138 verified claims, mostly from popular fact-checking websites, and identified 9.4K relevant tweets to those claims. Tweets were manually-annotated by veracity to support research on misinformation detection, which is one of the major problems faced during a pandemic. \ds{} supports two levels of misinformation detection over Twitter: verifying free-text claims (called \emph{claim-level verification}) and verifying claims expressed in tweets (called \emph{tweet-level verification}). Our dataset covers, in addition to health, claims related to other topical categories that were influenced by COVID-19, namely, social, politics, sports, entertainment, and religious. Moreover, we present benchmarking results for tweet-level verification on the dataset. We experimented with SOTA models of versatile approaches that either exploit content, user profiles features, temporal features and propagation structure of the conversational threads for tweet verification. 
\end{abstract}
\section{Introduction}
\label{intro}

\begin{table*}[h]
    \centering
\adjustbox{max width=\textwidth}{\begin{tabular}{l r r l c c c c}\toprule
{\bf Dataset} & {\bf \# Tweets} & {\bf \# Annotated} & {\bf Labels} & {\bf Multi-task} & {\bf Conversations} & {\bf Manual Annot.} \\\midrule
{\bf \citet{elhadad2020covid}} &       220K &       220K & misleading, real &          - &          - &          - \\
{\bf \citet{rayson2020covid}} &         1M &         2K & false, true, unrelated &          - &          - & \checkmark \\
{\bf \citet{mubarak2020}} &        30M &         8K & rumor, info, advice, … &          - &          - &   \checkmark      \\
{\bf \citet{alqurashi2021eating}} &       4.5M &       8.8K & misinformation, other &          - &          - & \checkmark \\
{\bf \ds{}} &       1M &       9.4K & false, true, other & \checkmark & \checkmark & \checkmark \\ \bottomrule
\end{tabular}}  
    \caption{Comparison between \ds{} and existing Arabic COVID-19 datasets for verification over tweets.
    }
    \label{data_compare}
\end{table*}

In addition to being a medium for the spread and consumption of news, Twitter has been shown to capture the dynamics of real-world events including the spread of diseases such as the seasonal influenza~\cite{kagashe2017} or more severe epidemics like Ebola~\cite{roy2020ebola}.
Since the first reported case of the Novel Coronavirus (COVID-19) in China, in November 2019, the COVID-19 topic has drawn the interest of many Arab users over Twitter. Their interest, reflected in the Arabic content on the platform, has reached a peak after two months when the first case was reported in the United Arab Emirates late in January 2020. This ongoing pandemic has, unsurprisingly, spiked discussions on Twitter covering a wide range of topics such as general information about the disease, preventive measures, procedures and newly-enforced decisions by governments, up-to-date statistics of the spread in the world, and even the change in our daily habits and work styles. With the great importance and spread of COVID-19 information, misinformation and fake news have infected the Twitter stream. An early study quantifying COVID-19 medical misinformation on Twitter found that 25\% of collected tweets contained misinformation~\cite{kouzy2020coronavirus}. During COVID-19 pandemic, we observed that misinformation stretched beyond spreading fake and potentially-harmful medical information, to information that can have adverse negative political effects (example: ``\emph{In light of the unresponsiveness of Yemeni government to requests of evacuation from Yemeni students in Wuhan, Sultan of Oman orders their evacuation}.") and economical effects too (example: ``\emph{Kuwaitis boycott AlMarai Saudi dairy company after reports on Coronavirus infected employees.}"). Combating the spread of such claims and verifying them becomes essential during this sensitive time. 

In this work, we aim to facilitate research on misinformation detection on social media during this complex and historical period of our time by introducing a manually-annotated Arabic dataset, \textbf{\ds}, that covers tweets spreading COVID-19 related claims. To construct \ds, we start from an existing COVID-19 Arabic dataset~\cite{haouari2020arcov19}, ArCOV-19, that is the first Arabic COVID-19 Twitter dataset with propagation networks. Our proposed \ds{} includes a set of 138 COVID-19 verified claims and 9.4K corresponding relevant tweets that were \textit{manually-annotated} to support both \emph{claim-level} and \emph{tweet-level} verification tasks. Claim-level verification is defined as follows: \emph{given a short free-text claim (in 1 or 2 sentences) and its corresponding relevant tweets, predict whether the claim is true or false}. Tweet-level verification is defined as follows: \emph{given a tweet containing a claim, detect whether it is true or false}.

To our knowledge, \ds{} is the only Arabic dataset made available to support both claim-level and tweet-level verification tasks in Twitter given the propagation networks of the tweets in general and on COVID-19 in particular. Some related Twitter datasets were recently released by \citet{alqurashi2021eating}, \citet{mubarak2020}, \citet{rayson2020covid}, and \citet{elhadad2020covid}. None of these datasets has the propagation networks of the tweets, and they either support the tweet verification task~\cite{elhadad2020covid,rayson2020covid}, misinformation detection~\cite{alqurashi2021eating} or multi-class categorization including rumors as one category~\cite{mubarak2020}. Differently from \ds{}, \citet{elhadad2020covid} used an automatic approach to annotate tweets, while \citet{rayson2020covid} only cover COVID-19 health-oriented claims. Starting from some health misinformation reported by the Ministry of Health in Saudi Arabia and the World Health Organization (WHO), ~\citet{alqurashi2021eating} annotated COVID-19 tweets as misinformation or not. Differently, our dataset covers in addition to health, other types of claims that were influenced by COVID-19, namely, social, politics, sports, entertainment, and religious. Table~\ref{data_compare} demonstrates the differences between \ds{} and the aforementioned datasets.

The contribution of this paper is three-fold:

\begin{itemize}
    \item We construct and release\footnote{\url{https://gitlab.com/bigirqu/ArCOV-19/-/tree/master/ArCOV19-Rumors}} the first Arabic dataset for misinformation detection over Twitter, covering both claim and tweet verification tasks. It contains 138 COVID-19 verified claims that scale to 9.4K labeled relevant tweets along with their propagation networks.
    \item We suggest and motivate several research tasks that can be addressed using our \textit{labeled} dataset for misinformation detection.
    \item We present benchmark results on tweet-level verification using SOTA models that either exploit content, user profiles, or the temporal features and the propagation structure of the conversational threads. Results offer baselines for future research on the problem. 
\end{itemize}

The remainder of the paper is organized as follows. 
We present studies related to COVID-19 misinformation analysis and datasets in Section~\ref{rw}. 
The construction of \ds{} is presented in Section~\ref{claims}. 
Several use cases supported by our dataset are discussed in Section~\ref{claim_use_cases}. Our benchmarks and their performances are presented in Section~\ref{benchmarks}. We detail the released components of the dataset in Section~\ref{sec:release}, and conclude in Section~\ref{conc}.

\section{Related Work} 

\label{rw}
The negative effects and spread of misinformation about COVID-19, triggered efforts to understand this phenomenon. Many studies analyzed misinformation spreading on Twitter related to COVID-19~\cite{shahi2020exploratory,cinelli2020covid,gallotti2020assessing,kouzy2020coronavirus,singh2020first}. \citet{singh2020first} analyzed the spread of five health misinformation over time.  \citet{shahi2020exploratory} analyzed false and partially false tweets that have been previously fact-checked by a fact-checking platform.  \citet{kouzy2020coronavirus} focused only on health misinformation and analyzed tweets negating health information by trusted sources (e.g., WHO). \citet{cui2020coaid} and \citet{hossain2020detecting} published an English dataset for claim verification,  and tweet verification respectively, while \citet{dharawat2020drink} released a Twitter dataset for health risk assessment of COVID-19-related English tweets.
 
 In this work, we extend an existing raw \textit{Arabic} COVID-19 Twitter dataset, \dscore{}~\cite{haouari2020arcov19}, to include manually-annotated tweets to support both claim and tweet verification. 
 To our knowledge, there is no work that released a large manually-annotated Arabic dataset for misinformation detection on COVID-19 at both claim and tweet levels. The closest work to ours is that done by \citet{elhadad2020covid} and \citet{rayson2020covid}. However, neither have the propagation networks of the annotated tweets, and they both solely support the tweet verification task. Moreover, \ds{} has the largest number of \emph{manually} annotated tweets.
 
Claim verification is a widely studied problem in general. Several non-COVID-19 datasets exist for claim verification (e.g., \cite{gorrell2019semeval}), however, they either support English language~\cite{zubiaga2016analysing,derczynski-etal-2017-semeval,gorrell2019semeval,ma2017detect} or Chinese language~\cite{ma2017detect}. Among these studies, some only focused on tweet-level verification~\cite{ma2017detect,liu2018early,bian2020rumor,khoo2020interpretable} while others  addressed claim-level verification~\cite{vosoughi2017rumor,dang2019early}.

There are few initiatives targeting Arabic online content, but they support rumors detection (i.e., rumor or non-rumor) \cite{alzanin2019rumor,alkhair2019arabic}, tweet verification without propagation networks \cite{elhadad2020covid, rayson2020covid}, misinformation detection~\cite{alqurashi2021eating}, or tweet credibility \cite{el2017cat}. Another notable work is a shared task by CLEF CheckThat! Lab for claim verification~\cite{elsayed2019overview,barron2020checkthat,barron2020overview}. To the best of our knowledge, this is the first Arabic dataset that supports both claim and tweet verification on social media using the propagation networks.

\section{Methodology} 
\label{claims}
We extended \dscore{}, an Arabic Twitter dataset about COVID-19~\cite{haouari2020arcov19}, by annotating a subset of the tweets to support research on misinformation detection, which is one of the major problems faced during a pandemic. We aim to support two classes of the misinformation detection problem (with variants) over Twitter: verifying free-text claims (called \emph{claim-level verification}) and verifying claims expressed in tweets (called \emph{tweet-level verification}), covering two common use cases. To that end, we need to collect COVID-19 \emph{verified} claims, then, for each claim, we need to find, in \dscore, corresponding \emph{relevant} tweets, and finally, among the relevant tweets, identify those that are either expressing the claim or negating it (to easily \emph{propagate} the veracity of the claims to the tweets). This section details how that pipeline was implemented (Sections \ref{sec:collect}-\ref{sec:network}). 

\subsection{Collecting COVID-19 Verified Claims}\label{sec:collect}
We manually collected a set of verified COVID-related claims from two popular Arabic fact-checking platforms: Fatabyyano\footnote{\url{https://fatabyyano.net}} and Misbar.\footnote{\url{https://misbar.com/}} We identified 113 false claims and 18 true claims during the period of \ds. To improve balance between false and true claims, we collect an additional 31  true claims from these sources:
\begin{enumerate}
    \item \textbf{Authoritative Health Organizations}: We retrieved, from \dscore{}, all tweets from Twitter accounts of the World Health Organization (WHO), ministries of public health in Arab countries, UNESCO, and UNICEF. Then, we manually extracted a set of true claims from the retrieved tweets. 
    \item \textbf{English Fact-checking Platforms}: We collected (and translated to Arabic) a few true claims from English Fact-checking platforms: PolitiFact,\footnote{\url{https://www.politifact.com/}}  Snopes,\footnote{\url{https://www.snopes.com/}} and FullFact.\footnote{\url{https://fullfact.org/}}
\end{enumerate}
That yielded a total of 162 claims, 113 false and 49 true, that are potentially suitable for our purpose.
\subsection{Finding Relevant Tweets}
For each claim, we manually constructed and refined a set of search queries in order to retrieve potentially-relevant tweets from \dscore{} using Boolean search. To help identify the queries and maximize our coverage, we considered keywords that appeared in the social media posts, expressing the claim, provided as examples by the fact-checking platforms, and we also interactively search in Twitter to find the best possible keywords that can retrieve relevant tweets to the claim. We then used the queries to search \dscore.
An example is the true claim ``Trump suggests injecting disinfectants as a treatment for COVID-19''.\footnote{\url{https://tinyurl.com/y4btakuw}} We noticed that the claim and also the tweet given as an example by Misbar fact-checking platform are using two different Arabic words that reflect ``disinfectants", namely \textit{mutaharat} (\textit{disinfectants} in English) and \textit{mueaqamat} (\textit{sterilizers} in English), so we used both in combination with either \textit{Trump} or \textit{US president} as search queries.

For each claim, we manually filtered the retrieved tweets to discard the non-relevant ones. This is conducted by one author of this paper. Examples of tweets that were considered relevant are tweets expressing or negating the claim, tweets having multiple claims including the target claim, tweets expressing advice, questions, or personal opinions about the claim, or even sarcastic tweets about it. Overall, more than half of the retrieved tweets were non-relevant, 
yielding 14,472 tweets relevant to the collected claims.

\subsection{Filtering Claims}
After identifying the relevant tweets, we excluded some claims for different reasons. We excluded claims with less than two relevant tweets due to extremely-insufficient content. 
We also excluded the ones (especially health-related claims) for which the veracity changed within the period of the dataset (e.g., ``\emph{Pets do not catch COVID-19}''), or those for which there are still no clear evidence with or against the claim (e.g., ``\emph{Bats are the source of COVID-19}''). Moreover, we discarded the claims that are not very specific, i.e., too general (e.g., ``\emph{Many Arab countries took security measures against those who refuse to quarantine}''). We eventually kept 95 false and 43 true claims, a total of 138 claims. Those claims have a total of 9,414 relevant tweets. The final set of claims is diverse and claims fall under different categories, as the misinformation propagating in Twitter during COVID-19 pandemic was not restricted to health. In fact, only 45 of them were health-related. The rest are distributed over social (38), political (22), religious (18), entertainment (9), and sports (6) topical categories.

\subsection{Annotating Relevant Tweets}
After collecting relevant tweets for each claim, it was time to identify tweets expressing or negating the claim, so that we can \emph{propagate} the label of the claim to them. For each claim, one of the authors of this paper labelled all of its relevant tweets by stance using the following three categories:\footnote{Full annotation guidelines:  \url{https://gitlab.com/bigirqu/ArCOV-19/-/blob/master/ArCOV19-Rumors/annotations_guidelines.md}}
\begin{itemize}
    \item ``Expressing same claim'' if the main (focused) claim in the tweet is restating, expressing, or rephrasing the target claim. This tweet then receives the \emph{same}  veracity of the target claim; thus inheriting its label (whether \emph{true} or \emph{false}).
    \item ``Negating the claim'' if the main (focused) claim in the tweet is negating or denying the target claim. The veracity of this tweet is then the \emph{opposite} of the veracity of the target claim, i.e., it is labeled as \emph{true} if the target claim is \emph{false}, and vice-versa.
    \item ``Other'' if the tweet cannot be labelled as one of the two earlier cases, e.g., expressing opinion or giving advice regarding the claim. 

\end{itemize}

\begin{figure}[!htb]
\centering
  \includegraphics[scale=0.42]{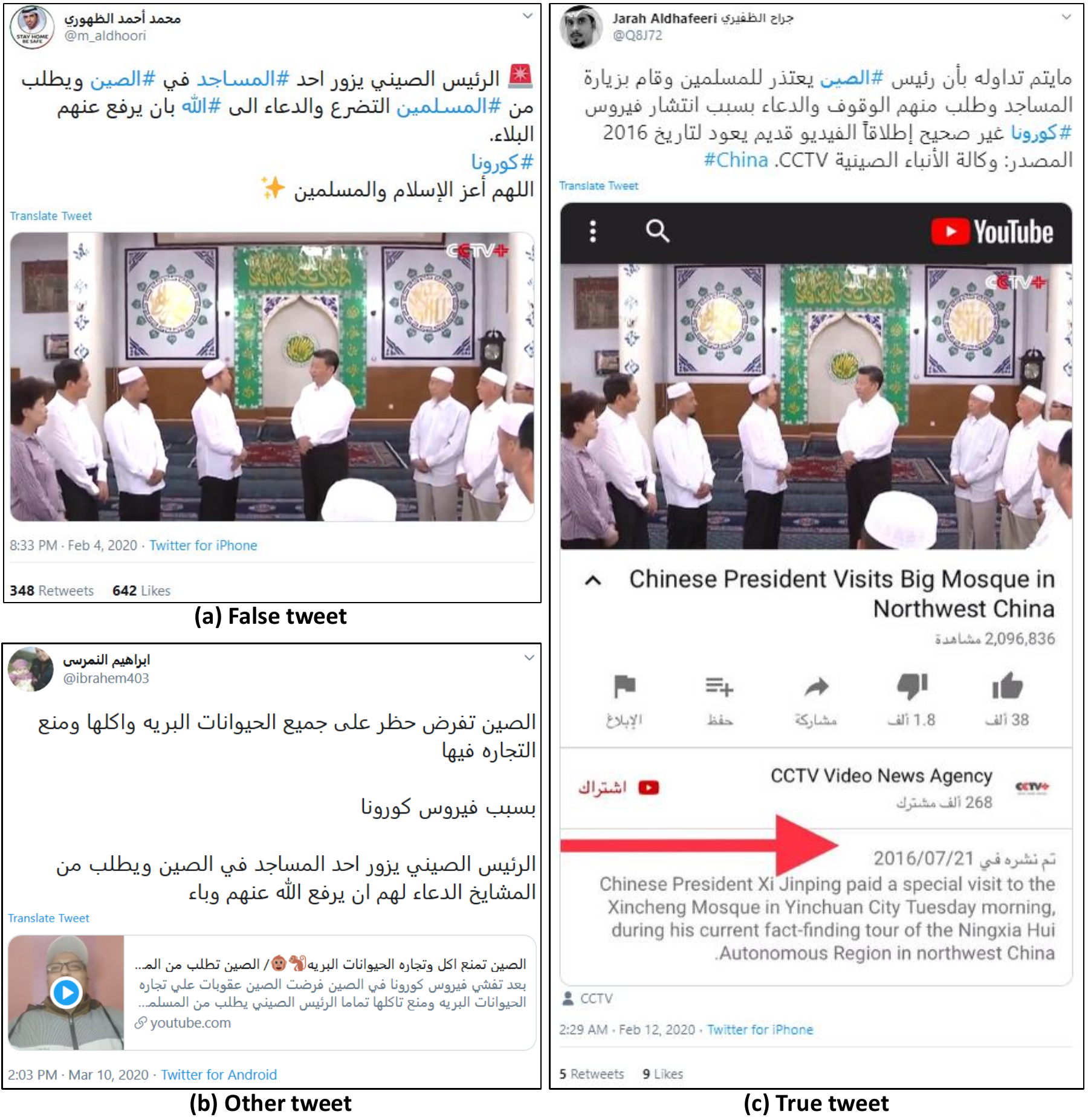}
  \caption{Example relevant tweets to a false claim:  ``\emph{The Chinese President visits one of the mosques in China and asks Muslims to pray to God to alleviate the pandemic}.''}

 \label{fig:tweetExample}
\end{figure}

Figure~\ref{fig:tweetExample} illustrates three example relevant tweets of a \emph{false} claim translated as ``\emph{The Chinese President visits one of the mosques in China and asks Muslims to pray to God to alleviate the pandemic}'', which is literally stated in the tweet shown in Figure~\ref{fig:tweetExample}(a), thus it is \emph{false}. 
On the contrary, the tweet illustrated in Figure~\ref{fig:tweetExample}(c) is denying the claim, and thus is \emph{true}. It is translated as ``\emph{What is being circulated about the Chinese President, apologizing to Muslims and visits the mosques and asking them to pray because of the spread of the \#Corona virus, is totally incorrect. The video is old and dated back to 2016. Source: the Chinese news agency CCTV}''. Finally, the tweet illustrated in Figure \ref{fig:tweetExample}(b) is labelled as ``\emph{other}'' due to expressing multiple claims. It is translated as ``\emph{China bans eating all wild animals and their trade due to Corona Virus. The Chinese President visits a mosque in China and asks the scholars to pray to alleviate the pandemic}''.
    
It is worth mentioning that expressing or negating the claim can be over an external link in the tweet, stated in an image, or said in a video, rather than in the text of the tweet. This was considered while annotating tweets. Providing such tweets in \ds{} allows the development of multi-modal systems, that can use signals in text, images, or videos, to make verification decisions.

\subsection{Data Quality}\label{sec:quality}
As mentioned earlier, our dataset was annotated by a single annotator to reduce annotation time and because the annotator was well-experienced with the task. To measure annotation quality, we randomly selected 10\% of the relevant tweets, and asked a second annotator to label them. We found that the agreement ratio between annotators is 0.87 and 0.80 for relevance and stance respectively. Due to subjectivity of the stance labelling task specifically (and since the annotator was also asked to consider images, videos, links, etc. in annotation which might lead to further subjectivity), we believe the agreement level is acceptable. 

\subsection{Collecting Propagation Networks}\label{sec:network}
\begin{figure}[!tb]
\centering
  \includegraphics[scale=0.32]{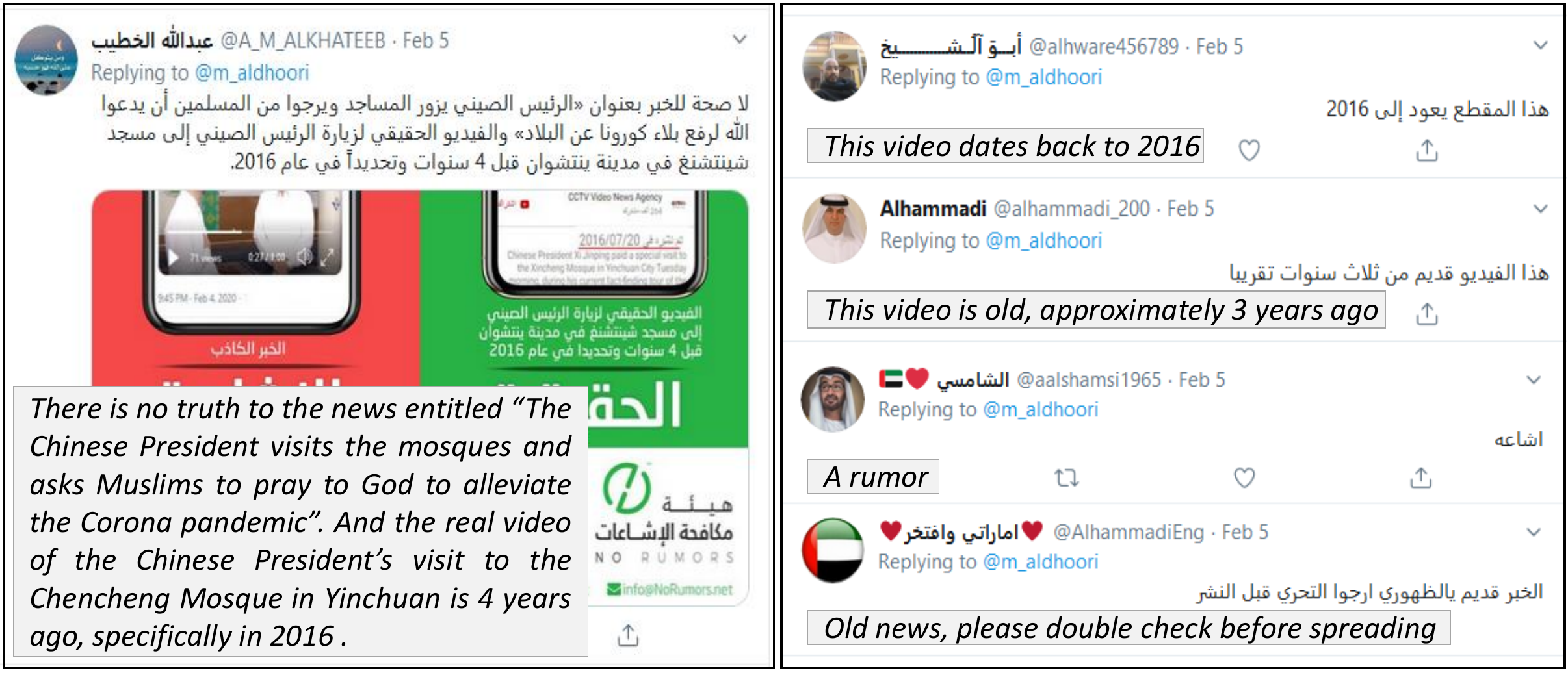}
  \caption{Evidence from replies (with translation) against the false tweet in Fig.~\ref{fig:tweetExample}(a).}
 \label{fig:repliesExample}
\end{figure}

We also collected the propagation networks (i.e., retweets and conversational threads) for each relevant tweet. The propagation networks for tweets that contain misinformation are essential to study its spreading behaviour and can constitute evidential signals for verification. Figure~\ref{fig:repliesExample} shows some replies to the false tweet presented in Figure~\ref{fig:tweetExample}(a). We notice that some replies have a clear stance against the claim. Moreover, one reply presents evidence that the claim in the original tweet is false. In addition to the replies, exploiting profiles of propagators can play a significant role in verifying the tweet~\cite{liu2018early}. 

\section{Immediate Use Cases}
\label{claim_use_cases}

\begin{table}
\centering
\caption{Statistics of Claims in \ds{} (RPs=Replies, RTs=Retweets).\label{tab:claimsstatistics}}
\adjustbox{max width=\columnwidth}{
    \begin{tabular}{lrrrr}
    \toprule
         &\multicolumn{2}{r}{\textbf{Claims Subset}} & \multicolumn{2}{r}{\textbf{Tweet Vf. Subset}} \\
    \midrule
        \textbf{Total tweets} & 9,414 & & 3,584&\\
        \,\,\,\,\, False Tweets & 1,753& (18.6\%) & 1,753 &(48.9\%) \\ 
        \,\,\,\,\, True Tweets & 1,831& (19.4\%) & 1,831 &(51.1\%) \\
        \,\,\,\,\, Other Tweets &5,830 &(61.9\%)& 0&   \\
    \midrule
        Average tweets / claim              &68 && 26&\\ 
        Tweets with RPs & 3,161 &(33.6\%)& 1,222 &(34.1\%)\\
        Tweets with RTs &  3,810&(40.5\%)& 1,629 &(45.5\%)\\
        Tweets with RPs \& RTs    & 2,006 &(21.3\%)& 772 &(21.5\%)\\
        Average RPs / tweet             &33 && 31&\\
        Average RTs per / tweet            & 16&& 19&\\ 
    \bottomrule
    \end{tabular}
}
\end{table}

Table~\ref{tab:claimsstatistics} presents a statistical summary of the labeled tweets in \ds. We define two subsets of tweets. The first is denoted as \textbf{Claims Subset}, which includes all relevant tweets of the claims (labeled as \emph{true}, \emph{false}, or \emph{other}). This is of high interest since all tweets relevant to a specific claim can be used in verifying it. 
The other is denoted as \textbf{Tweet Verification Subset}, which only includes the relevant tweets that are either \emph{expressing} or \emph{denying} the original claims, excluding the ones labeled as \textit{other}. This is also of high interest since each of those tweets is subject to verification.  

The table indicates that, out of 9.4k labelled relevant tweets, about 3.6k of them are either true or false (constituting the tweet verification subset); each is considered a separate tweet-level verification query. We notice that the distribution of \emph{true} vs. \emph{false} tweets is balanced, making it a good resource for training verification systems. We also notice that a good portion of both subsets have retweets and replies, indicating potentially-useful propagation networks. Accordingly and based on the two subsets, our labeled data can support three different misinformation detection tasks.

\subsection{Claim-level Verification}
This task is defined as follows: \emph{given a claim and all corresponding relevant tweets }(\emph{with their propagation networks}), \emph{detect the veracity of the claim, i.e., whether the claim is \textit{true} or \textit{false}}. There are some initiatives to support Arabic claim verification. However, in one of the most prominent ones (CheckThat! lab at CLEF-2019), a pre-defined set of Web page was used in verification~\cite{elsayed2019overview} while in \ds{} we focus on Twitter.
\subsection{Tweet-level Verification}\label{tweetlevelverif}
This task is defined as follows: \emph{given a tweet} (\emph{with its propagation networks}), \emph{detect its veracity, i.e., whether the tweet is \textit{true} or \textit{false}}. Addressing this task in Arabic has never been studied. Existing studies for Arabic tweet verification mainly rely on the source tweet content only \cite{elhadad2020ensemble,rayson2020covid}, or, additionally, the potentially-relevant Web pages~\cite{barron2020checkthat,barron2020overview}.

\ds{} supports another variant of this task. This variant makes also available (to the verification system) the tweets that are relevant to its target claim but were posted \emph{earlier} (with their propagation networks) allowing for early claim verification. To our knowledge, there is no study that addressed this problem. This is an interesting problem for several reasons. First, with the lack of any propagation networks for a target tweet, a verification system can still exploit networks of earlier relevant tweets to verify the tweet. Second, as time is critical to debunk fake claims, exploiting relevant tweets posted earlier allows verifying the tweet as soon as it is posted, without waiting for its propagation networks. Third, even if the target tweet has propagation networks, relevant tweets might provide more evidence, hence improving verification accuracy. 

\subsection{Claim Retrieval}\label{tweetretrieval}

This task is defined as follows: \emph{given a tweet that expresses a claim} (\emph{i.e., a tweet in the tweet verification subset}), \emph{find all tweets that are expressing the same claim}. To our knowledge, this task is under-studied; an exception is the work done by \citet{shaar2020known} and the task proposed by~\citet{barron2020checkthat} in CLEF CheckThat! 2020 lab;~\footnote{\url{https://sites.google.com/view/clef2020-checkthat/}} however, they focus more on claims than tweets and they release English-only datasets to support this task. 
Solving this problem helps in applications like finding previously-verified tweets, or clustering tweets expressing the same claim, to avoid re-verification. 
\section{Experiments and Evaluation}\label{benchmarks}
In this section, we present \emph{benchmarking results} using SOTA models on the \emph{tweet-level} verification task to facilitate future research. We experimented with variant models that either exploit content only, user profile features, temporal features, or propagation structure of the conversational threads for tweet verification. 
We present our preprocessing approach in Section~\ref{preprocessing}, the SOTA models used for benchmarking in Section~\ref{models}, and finally a discussion of the results in Section~\ref{results}.

\subsection{Preprocessing} \label{preprocessing}
In all experiments, we used our \textbf{tweets with RPs} subset, since some of the SOTA methods we experiment with depend on replies of tweets to be verified (\emph{target tweets} hereafter). In \ds{}, some of the target tweets were replies to previous tweets. 
In our experiments, we considered the direct and indirect replies that were posted \emph{only after} the target tweet. 
We also eliminated target tweets that have no textual replies, i.e., containing only emojis, non Arabic content, images, or videos.
We ended up with 1,108 tweets, 597 True and 511 False, with 16 replies on average. We split our data by target tweets over 5 folds, ensuring there is no overlap of claims across folds (i.e., no overlap between tweets of the same claim).\footnote{Folds used in our experiments:  \url{https://gitlab.com/bigirqu/ArCOV-19/-/tree/master/ArCOV19-Rumors/baselines_folds}}

We processed the tweets and the replies by removing non-Arabic letters, URLs, handles, special characters, and diacritics. 

\subsection{Verification Models}\label{models}
We experimented with two SOTA models that exploit the propagation networks for tweet verification. Moreover, due to proven effectiveness of BERT-based classifiers in versatile text classification tasks, we elect to develop a simple and effective BERT-based classifier that we fine-tune for the task. In this BERT-based classification architecture, we experimented with two Arabic pre-trained BERT models, namely, AraBERT and MARBERT. Further details on the models are presented below.
\begin{enumerate}
    \item Bi-GCN~\citet{bian2020rumor}: a bidirectional Graph Convolutional Networks model that leverages Graph Convolutional Networks to verify tweets given the target tweet and replies content, in addition to the replies tree structure. The model was called bidirectional for its ability to capture the bottom-up and top-down replies tree for each tweet. We used the authors implementation and setup\footnote{\url{https://github.com/TianBian95/BiGCN}} in our experiments. Each tweet/reply was represented using a vector of 5K content features, which are the most frequent 5K words in the dataset excluding stop words. 
    \item PPC-RNN+CNN~\citet{liu2018early}: a multivariate time series model that exploits user profiles to verify tweets. Each time step involves a single user represented by a vector of user features.  
   A user can be the author of the target tweet or a reply to it. The user vectors were sorted by posting time starting from the target tweet to capture the temporal features. In the original paper, the model uses a fixed number of time steps. In our experiments, we find the average number of replies per target tweet is 16, thus, we set the time steps to 17 taking into account the target tweet. We implemented this model using Keras\footnote{\url{https://keras.io/}} following the same setup presented in the original paper. 
    
    \item AraBERT~\citet{baly2020arabert}: a pretrained BERT model on a
     large-scale Arabic corpus of news articles. We fine-tuned the model to classify the tweets as True or False \emph{given the target tweet content only}. We used the fine-tuning implementation with the same hyperparameters setup shared by the model developers,\footnote{\url{https://colab.research.google.com/drive/1P9iQHtUH5KUbTVtp8B4-AopZzEEPE0lw?usp=sharing}} using AraBERTv1, which is based on pre-segmentation using Farasa segmenter,\footnote{\url{https://farasa.qcri.org/}} and we set the max sequence length to 128. The model was fine-tuned with a batch size of 16 for 8 epochs, with a learning rate of $10^{-5}$.
     
    \item MARBERT~\citet{abdul2020arbert}: a pretrained BERT model on a large Twitter dataset. We selected this model since in our task, we are solely working with Twitter data and thus, a pretrained model from the same domain might be more suitable. We fine-tuned the model for a classification task given the same input as AraBERT. We used the authors' code\footnote{\url{https://colab.research.google.com/drive/1M0ls7EPUi1dwqIDh6HNfJ5y826XvcgGX?usp=sharing}} and setup to fine-tune the model. The model was fine-tuned with a batch size of 32 for 5 epochs, with a learning rate of $2*10^{-6}$. Similar to AraBERT, the max sequence length was set to 128.
\end{enumerate}

\begin{table}
\centering
\caption{Tweet-level verification results.}
\adjustbox{max width=\columnwidth}{
    \begin{tabular}{l l r r r r r}
    \toprule
    Model   & Class  & \multicolumn{1}{r}{Acc}   & \multicolumn{1}{r}{P}    & \multicolumn{1}{r}{R}   & \multicolumn{1}{r}{F1}& \multicolumn{1}{r}{macro-F1}    \\ \midrule
    Majority                                &   -    & \multicolumn{1}{r}{0.533} & \multicolumn{1}{r}{0.266} & \multicolumn{1}{r}{0.500} & -& \multicolumn{1}{r}{0.343}  \\ \hline
    \multirow{2}{*}{Bi-GCN}                            & F     & \multirow{2}{*}{0.669}     & 0.731                      & 0.526                    & 0.580 & \multirow{2}{*}{0.649}                    \\ 
                                                       & T     &                            & 0.662                      & 0.830                    & 0.718                      \\ 
    \multicolumn{1}{l}{\multirow{2}{*}{RNN+CNN}} & F     & \multirow{2}{*}{0.682}     & 0.631                      & 0.783                    & 0.688 &\multirow{2}{*}{0.673}                     \\ 
    \multicolumn{1}{l}{}                             & T     &                            & 0.749                      & 0.600                    & 0.657                      \\ \hline
     \multicolumn{1}{l}{\multirow{2}{*}{AraBERT}}                          & F     & \multirow{2}{*}{0.730}     & 0.733                      & 0.671                    & 0.691   &\multirow{2}{*}{0.713}                    \\ 
                                                   \multicolumn{1}{l}{}       & T     &                            & 0.719                      & 0.763                    & 0.735                    \\ 
    \multirow{2}{*}{MARBERT}                           & F     & \multirow{2}{*}{0.757}     & 0.763                      & 0.692                    & 0.717     &\multirow{2}{*}{0.740}                  \\ 
                                                       & T     &                            & 0.743                      & 0.788                    & 0.762                      \\ \bottomrule 
    \end{tabular}
}
\label{results-tab}
\end{table}
\subsection{Results and Discussion}\label{results}
Table~\ref{results-tab} presents the performance of the four models, in addition to a simple majority baseline. For each model, we report the overall accuracy, and also F1, precision, and recall for each class.  
Starting from the same set of 5 folds, we trained each model 5 times, each with a random seed, and we report the average over those runs then over folds.\footnote{It is worth noting that since we set the hyper-parameters values as reported by original papers, better performance might be achieved if hyper-parameters are tuned on our dataset.}

Results show that all models significantly outperform the majority baseline. Moreover, the pre-trained BERT models are superior to the other models, despite the fact that they leverage the reply tree in addition to the target tweet. More specifically, MARBERT exhibits better performance than AraBERT, indicating the effectiveness of matching the pre-training domain with the testing one, and achieving 0.74 macro-averaged F1. 

The results also demonstrate that the models are better in detecting True claims than False claims, with an exception of the PPC-RNN+CNN model. This can be attributed to the fact that the average number of replies for False tweets is 12 vs. 19 for True tweets. We note that, in case the time steps are actually less than 17 (the average indicated earlier), the model \emph{randomly} fills the missing time steps with user features from other repliers to the target tweet. Such limitation in the model might lead to added noise and thus a poor performance with False tweets.

Moreover, the Bi-GCN model is language-dependent, and since we are working with Arabic data, we may need to test preprocessing the data with different techniques, such as considering stemming the content or replacing the top 5000 words features with content embeddings instead.
\section{Data Release}
\label{sec:release}
In summary, we release the following resources as \ds{} dataset, taking into consideration Twitter content redistribution policy:\footnote{\url{https://developer.twitter.com/en/developer-terms/agreement-and-policy}}
\setlist{nolistsep}
\begin{itemize}[noitemsep]
\itemsep0em 
    \item \textbf{Verified Claims}: 138 verified claims, each labeled as true or false.
    \item \textbf{Claims Subset}: IDs of the tweets relevant to the verified claims, each labeled as true, false, or other.
    \item \textbf{Propagation Networks of the Claims Subsets}: which includes for each tweet in the claim subset:
        \begin{itemize}
            \item \textbf{Retweets}: IDs of the full retweet set.
            \item \textbf{Conversational Threads}: tweet IDs of the full reply thread (including direct and indirect replies).
        \end{itemize}
    \item \textbf{Annotation guidelines:} guidelines used to annotate the relevant tweets. 
    \item \textbf{Baselines folds:} tweet IDs for folds used to train our baselines.
\end{itemize}
Along with the dataset, we plan to provide some pointers to publicly-available crawlers that users can easily use to crawl the tweets given their IDs.
\section{Conclusion}
In this paper, we presented \ds{}, which is an Arabic Twitter dataset that supports both claim-level and tweet-level verification tasks given the propagation networks. We released 138 verified claims associated with 9.4K relevant tweets. Our dataset covers, in addition to health, other types of claims that were influenced by COVID-19, namely, social, politics, sports, entertainment, and religious.
To facilitate future research, we presented benchmarking performance results of some SOTA models on the tweet verification task. 
\label{conc}

\section*{Acknowledgments}
The work of Tamer Elsayed and Maram Hasanain was made possible by NPRP grant\# NPRP 11S-1204-170060 from the Qatar National Research Fund (a member of Qatar Foundation). The work of Reem Suwaileh was supported by GSRA grant\# GSRA5-1-0527-18082 from the Qatar National Research Fund and the work of Fatima Haouari was supported by GSRA grant\# GSRA6-1-0611-19074 from the Qatar National Research Fund. The statements made herein are solely the responsibility of the authors. 

\bibliography{0.arcov19-main}

\begin{thebibliography}{34}
\expandafter\ifx\csname natexlab\endcsname\relax\def\natexlab#1{#1}\fi

\bibitem[{Abdul-Mageed et~al.(2020)Abdul-Mageed, Elmadany, and
  Nagoudi}]{abdul2020arbert}
Muhammad Abdul-Mageed, AbdelRahim Elmadany, and El~Moatez~Billah Nagoudi. 2020.
\newblock Arbert \& marbert: Deep bidirectional transformers for arabic.
\newblock \emph{arXiv preprint arXiv:2101.01785}.

\bibitem[{Alkhair et~al.(2019)Alkhair, Meftouh, Sma{\"\i}li, and
  Othman}]{alkhair2019arabic}
Maysoon Alkhair, Karima Meftouh, Kamel Sma{\"\i}li, and Nouha Othman. 2019.
\newblock An arabic corpus of fake news: Collection, analysis and
  classification.
\newblock In \emph{International Conference on Arabic Language Processing},
  pages 292--302. Springer.

\bibitem[{Alqurashi et~al.(2021)Alqurashi, Hamoui, Alashaikh, Alhindi, and
  Alanazi}]{alqurashi2021eating}
Sarah Alqurashi, Btool Hamoui, Abdulaziz Alashaikh, Ahmad Alhindi, and Eisa
  Alanazi. 2021.
\newblock Eating garlic prevents covid-19 infection: Detecting misinformation
  on the arabic content of twitter.
\newblock \emph{arXiv preprint arXiv:2101.05626}.

\bibitem[{Alsudias and Rayson(2020)}]{rayson2020covid}
Lama Alsudias and Paul Rayson. 2020.
\newblock {COVID-19} and {Arabic} {Twitter}: How can {Arab} world governments
  and public health organizations learn from social media?
\newblock In \emph{Proceedings of the 1st Workshop on {NLP} for {COVID-19} at
  {ACL} 2020}. Association for Computational Linguistics.

\bibitem[{Alzanin and Azmi(2019)}]{alzanin2019rumor}
Samah~M Alzanin and Aqil~M Azmi. 2019.
\newblock Rumor detection in arabic tweets using semi-supervised and
  unsupervised expectation--maximization.
\newblock \emph{Knowledge-Based Systems}, 185:104945.

\bibitem[{Baly et~al.(2020)Baly, Hajj et~al.}]{baly2020arabert}
Fady Baly, Hazem Hajj, et~al. 2020.
\newblock Arabert: Transformer-based model for arabic language understanding.
\newblock In \emph{Proceedings of the 4th Workshop on Open-Source Arabic
  Corpora and Processing Tools, with a Shared Task on Offensive Language
  Detection}, pages 9--15.

\bibitem[{Barr{\'o}n-Cede{\~n}o et~al.(2020)Barr{\'o}n-Cede{\~n}o, Elsayed,
  Nakov, Da~San~Martino, Hasanain, Suwaileh, and Haouari}]{barron2020checkthat}
Alberto Barr{\'o}n-Cede{\~n}o, Tamer Elsayed, Preslav Nakov, Giovanni
  Da~San~Martino, Maram Hasanain, Reem Suwaileh, and Fatima Haouari. 2020.
\newblock Checkthat! at clef 2020: Enabling the automatic identification and
  verification of claims in social media.
\newblock In \emph{European Conference on Information Retrieval}, pages
  499--507. Springer.

\bibitem[{Barr{\'o}n-Cedeno et~al.(2020)Barr{\'o}n-Cedeno, Elsayed, Nakov,
  Da~San~Martino, Hasanain, Suwaileh, Haouari, Babulkov, Hamdan, Nikolov
  et~al.}]{barron2020overview}
Alberto Barr{\'o}n-Cedeno, Tamer Elsayed, Preslav Nakov, Giovanni
  Da~San~Martino, Maram Hasanain, Reem Suwaileh, Fatima Haouari, Nikolay
  Babulkov, Bayan Hamdan, Alex Nikolov, et~al. 2020.
\newblock Overview of checkthat! 2020: Automatic identification and
  verification of claims in social media.
\newblock In \emph{International Conference of the Cross-Language Evaluation
  Forum for European Languages}, pages 215--236. Springer.

\bibitem[{Bian et~al.(2020)Bian, Xiao, Xu, Zhao, Huang, Rong, and
  Huang}]{bian2020rumor}
Tian Bian, Xi~Xiao, Tingyang Xu, Peilin Zhao, Wenbing Huang, Yu~Rong, and
  Junzhou Huang. 2020.
\newblock Rumor detection on social media with bi-directional graph
  convolutional networks.
\newblock In \emph{Proceedings of the AAAI Conference on Artificial
  Intelligence}, pages 549--556.

\bibitem[{Cinelli et~al.(2020)Cinelli, Quattrociocchi, Galeazzi, Valensise,
  Brugnoli, Schmidt, Zola, Zollo, and Scala}]{cinelli2020covid}
Matteo Cinelli, Walter Quattrociocchi, Alessandro Galeazzi, Carlo~Michele
  Valensise, Emanuele Brugnoli, Ana~Lucia Schmidt, Paola Zola, Fabiana Zollo,
  and Antonio Scala. 2020.
\newblock The covid-19 social media infodemic.
\newblock \emph{arXiv preprint arXiv:2003.05004}.

\bibitem[{Cui and Lee(2020)}]{cui2020coaid}
Limeng Cui and Dongwon Lee. 2020.
\newblock Coaid: Covid-19 healthcare misinformation dataset.
\newblock \emph{arXiv preprint arXiv:2006.00885}.

\bibitem[{Dang et~al.(2019)Dang, Moh’d, Islam, and Milios}]{dang2019early}
Anh Dang, Abidalrahman Moh’d, Aminul Islam, and Evangelos Milios. 2019.
\newblock Early detection of rumor veracity in social media.
\newblock In \emph{Proceedings of the 52nd Hawaii International Conference on
  System Sciences}.

\bibitem[{Derczynski et~al.(2017)Derczynski, Bontcheva, Liakata, Procter, Wong
  Sak~Hoi, and Zubiaga}]{derczynski-etal-2017-semeval}
Leon Derczynski, Kalina Bontcheva, Maria Liakata, Rob Procter, Geraldine Wong
  Sak~Hoi, and Arkaitz Zubiaga. 2017.
\newblock {S}em{E}val-2017 task 8: {R}umour{E}val: Determining rumour veracity
  and support for rumours.
\newblock In \emph{Proceedings of the 11th International Workshop on Semantic
  Evaluation ({S}em{E}val-2017)}. Association for Computational Linguistics.

\bibitem[{Dharawat et~al.(2020)Dharawat, Lourentzou, Morales, and
  Zhai}]{dharawat2020drink}
Arkin Dharawat, Ismini Lourentzou, Alex Morales, and ChengXiang Zhai. 2020.
\newblock Drink bleach or do what now? covid-hera: A dataset for risk-informed
  health decision making in the presence of covid19 misinformation.
\newblock \emph{arXiv preprint arXiv:2010.08743}.

\bibitem[{El~Ballouli et~al.(2017)El~Ballouli, El-Hajj, Ghandour, Elbassuoni,
  Hajj, and Shaban}]{el2017cat}
Rim El~Ballouli, Wassim El-Hajj, Ahmad Ghandour, Shady Elbassuoni, Hazem Hajj,
  and Khaled Shaban. 2017.
\newblock Cat: Credibility analysis of arabic content on twitter.
\newblock In \emph{Proceedings of the Third Arabic Natural Language Processing
  Workshop}, pages 62--71.

\bibitem[{Elhadad et~al.(2020{\natexlab{a}})Elhadad, Li, and
  Gebali}]{elhadad2020covid}
Mohamed~K Elhadad, Kin~Fun Li, and Fayez Gebali. 2020{\natexlab{a}}.
\newblock Covid-19-fakes: a twitter (arabic/english) dataset for detecting
  misleading information on covid-19.
\newblock In \emph{International Conference on Intelligent Networking and
  Collaborative Systems}, pages 256--268. Springer.

\bibitem[{Elhadad et~al.(2020{\natexlab{b}})Elhadad, Li, and
  Gebali}]{elhadad2020ensemble}
Mohamed~K Elhadad, Kin~Fun Li, and Fayez Gebali. 2020{\natexlab{b}}.
\newblock An ensemble deep learning technique to detect covid-19 misleading
  information.
\newblock In \emph{International Conference on Network-Based Information
  Systems}, pages 163--175.

\bibitem[{Elsayed et~al.(2019)Elsayed, Nakov, Barr{\'o}n-Cede{\~n}o, Hasanain,
  Suwaileh, Da~San~Martino, and Atanasova}]{elsayed2019overview}
Tamer Elsayed, Preslav Nakov, Alberto Barr{\'o}n-Cede{\~n}o, Maram Hasanain,
  Reem Suwaileh, Giovanni Da~San~Martino, and Pepa Atanasova. 2019.
\newblock Overview of the clef-2019 checkthat! lab: Automatic identification
  and verification of claims.
\newblock In \emph{International Conference of the Cross-Language Evaluation
  Forum for European Languages}, pages 301--321. Springer.

\bibitem[{Gallotti et~al.(2020)Gallotti, Valle, Castaldo, Sacco, and
  De~Domenico}]{gallotti2020assessing}
Riccardo Gallotti, Francesco Valle, Nicola Castaldo, Pierluigi Sacco, and
  Manlio De~Domenico. 2020.
\newblock Assessing the risks of" infodemics" in response to covid-19
  epidemics.
\newblock \emph{arXiv preprint arXiv:2004.03997}.

\bibitem[{Gorrell et~al.(2019)Gorrell, Kochkina, Liakata, Aker, Zubiaga,
  Bontcheva, and Derczynski}]{gorrell2019semeval}
Genevieve Gorrell, Elena Kochkina, Maria Liakata, Ahmet Aker, Arkaitz Zubiaga,
  Kalina Bontcheva, and Leon Derczynski. 2019.
\newblock Semeval-2019 task 7: Rumoureval, determining rumour veracity and
  support for rumours.
\newblock In \emph{Proceedings of the 13th International Workshop on Semantic
  Evaluation}, pages 845--854.

\bibitem[{Haouari et~al.(2021)Haouari, Hasanain, Suwaileh, and
  Elsayed}]{haouari2020arcov19}
Fatima Haouari, Maram Hasanain, Reem Suwaileh, and Tamer Elsayed. 2021.
\newblock Arcov-19: The first arabic covid-19 twitter dataset with propagation
  networks.
\newblock In \emph{Proceedings of the Sixth Arabic Natural Language Processing
  Workshop}, WANLP 2021.

\bibitem[{Hossain et~al.(2020)Hossain, Logan~IV, Ugarte, Matsubara, Young, and
  Singh}]{hossain2020detecting}
Tamanna Hossain, Robert~L. Logan~IV, Arjuna Ugarte, Yoshitomo Matsubara, Sean
  Young, and Sameer Singh. 2020.
\newblock {COVIDL}ies: Detecting {COVID}-19 misinformation on social media.
\newblock In \emph{Proceedings of the 1st Workshop on {NLP} for {COVID}-19
  (Part 2) at {EMNLP} 2020}.

\bibitem[{Kagashe et~al.(2017)Kagashe, Yan, and Suheryani}]{kagashe2017}
Ireneus Kagashe, Zhijun Yan, and Imran Suheryani. 2017.
\newblock Enhancing seasonal influenza surveillance: topic analysis of widely
  used medicinal drugs using twitter data.
\newblock \emph{Journal of medical Internet research}, 19(9):e315.

\bibitem[{Khoo et~al.(2020)Khoo, Chieu, Qian, and
  Jiang}]{khoo2020interpretable}
Ling Min~Serena Khoo, Hai~Leong Chieu, Zhong Qian, and Jing Jiang. 2020.
\newblock Interpretable rumor detection in microblogs by attending to user
  interactions.
\newblock \emph{arXiv preprint arXiv:2001.10667}.

\bibitem[{Kouzy et~al.(2020)Kouzy, Abi~Jaoude, Kraitem, El~Alam, Karam, Adib,
  Zarka, Traboulsi, Akl, and Baddour}]{kouzy2020coronavirus}
Ramez Kouzy, Joseph Abi~Jaoude, Afif Kraitem, Molly~B El~Alam, Basil Karam,
  Elio Adib, Jabra Zarka, Cindy Traboulsi, Elie~W Akl, and Khalil Baddour.
  2020.
\newblock Coronavirus goes viral: quantifying the covid-19 misinformation
  epidemic on twitter.
\newblock \emph{Cureus}, 12(3).

\bibitem[{Liu and Wu(2018)}]{liu2018early}
Yang Liu and Yi-Fang~Brook Wu. 2018.
\newblock Early detection of fake news on social media through propagation path
  classification with recurrent and convolutional networks.
\newblock In \emph{Thirty-Second AAAI Conference on Artificial Intelligence}.

\bibitem[{Ma et~al.(2017)Ma, Gao, and Wong}]{ma2017detect}
Jing Ma, Wei Gao, and Kam-Fai Wong. 2017.
\newblock Detect rumors in microblog posts using propagation structure via
  kernel learning.
\newblock In \emph{Proceedings of the 55th Annual Meeting of the Association
  for Computational Linguistics (Volume 1: Long Papers)}, pages 708--717.

\bibitem[{Mubarak and Hassan(2020)}]{mubarak2020}
Hamdy Mubarak and Sabit Hassan. 2020.
\newblock Arcorona: Analyzing arabic tweets in the early days of coronavirus
  (covid-19) pandemic.
\newblock \emph{arXiv preprint arXiv:2012.01462}.

\bibitem[{Roy et~al.(2020)Roy, Moreau, Rousseau, Mercier, Wilson, and
  Atlani-Duault}]{roy2020ebola}
Melissa Roy, Nicolas Moreau, C{\'e}cile Rousseau, Arnaud Mercier, Andrew
  Wilson, and La{\"e}titia Atlani-Duault. 2020.
\newblock Ebola and localized blame on social media: analysis of twitter and
  facebook conversations during the 2014--2015 ebola epidemic.
\newblock \emph{Culture, Medicine, and Psychiatry}, 44(1):56--79.

\bibitem[{Shaar et~al.(2020)Shaar, Martino, Babulkov, and
  Nakov}]{shaar2020known}
Shaden Shaar, Giovanni Da~San Martino, Nikolay Babulkov, and Preslav Nakov.
  2020.
\newblock That is a known lie: Detecting previously fact-checked claims.
\newblock \emph{arXiv preprint arXiv:2005.06058}.

\bibitem[{Shahi et~al.(2020)Shahi, Dirkson, and
  Majchrzak}]{shahi2020exploratory}
Gautam~Kishore Shahi, Anne Dirkson, and Tim~A Majchrzak. 2020.
\newblock An exploratory study of covid-19 misinformation on twitter.
\newblock \emph{arXiv preprint arXiv:2005.05710}.

\bibitem[{Singh et~al.(2020)Singh, Bansal, Bode, Budak, Chi, Kawintiranon,
  Padden, Vanarsdall, Vraga, and Wang}]{singh2020first}
Lisa Singh, Shweta Bansal, Leticia Bode, Ceren Budak, Guangqing Chi, Kornraphop
  Kawintiranon, Colton Padden, Rebecca Vanarsdall, Emily Vraga, and Yanchen
  Wang. 2020.
\newblock A first look at covid-19 information and misinformation sharing on
  twitter.
\newblock \emph{arXiv preprint arXiv:2003.13907}.

\bibitem[{Vosoughi et~al.(2017)Vosoughi, Mohsenvand, and
  Roy}]{vosoughi2017rumor}
Soroush Vosoughi, Mostafa~‘Neo’ Mohsenvand, and Deb Roy. 2017.
\newblock Rumor gauge: Predicting the veracity of rumors on twitter.
\newblock \emph{ACM Transactions on Knowledge Discovery from Data (TKDD)},
  11(4):1--36.

\bibitem[{Zubiaga et~al.(2016)Zubiaga, Liakata, Procter, Hoi, and
  Tolmie}]{zubiaga2016analysing}
Arkaitz Zubiaga, Maria Liakata, Rob Procter, Geraldine Wong~Sak Hoi, and Peter
  Tolmie. 2016.
\newblock Analysing how people orient to and spread rumours in social media by
  looking at conversational threads.
\newblock \emph{PloS one}, 11(3).

\end{thebibliography}
\bibliographystyle{acl_natbib}

\end{document}